\documentclass[conference]{IEEEtran}
\IEEEoverridecommandlockouts
\usepackage{cite}
\usepackage{amsmath,amssymb,amsfonts}
\usepackage{algorithmic}
\usepackage{graphicx}
\usepackage{textcomp}
\usepackage{xcolor}
\usepackage{subcaption}
\usepackage{float}
\usepackage{multirow} 

\def\BibTeX{{\rm B\kern-.05em{\sc i\kern-.025em b}\kern-.08em
    T\kern-.1667em\lower.7ex\hbox{E}\kern-.125emX}}
\begin{document}

\title{Machine Learning-Based Prediction of Speech Arrest During Direct Cortical Stimulation Mapping}

\author{
Nikasadat Emami$^{1}$, Amirhossein Khalilian-Gourtani$^{2}$, Jianghao Qian$^{1,*}$,\\
Antoine Ratouchniak$^{1}$, Xupeng Chen$^{1}$, Yao Wang$^{1,3}$, Adeen Flinker$^{2,3}$ \\
$^{1}$Department of Electrical and Computer Engineering, New York University, New York, USA\\
$^{2}$Department of Neurology, New York University, New York, USA\\
$^{3}$Department of Biomedical Engineering, New York University, New York, USA\\
}

\maketitle

\renewcommand{\thefootnote}{\fnsymbol{footnote}}
\footnotetext[0]{\hspace{-1.8ex}%
© 2025 IEEE. Personal use of this material is permitted. Permission from IEEE must be obtained for all other uses, in any current or future media, including reprinting/republishing this material for advertising or promotional purposes, creating new collective works, for resale or redistribution to servers or lists, or reuse of any copyrighted component of this work in other works.}

\footnotetext[1]{\vspace{0.35\baselineskip}%
Visiting student from Tsinghua University during the time of this work.}
\renewcommand{\thefootnote}{\arabic{footnote}}

\begin{abstract}
Identifying cortical regions critical for speech is essential for safe brain surgery in or near language areas. While Electrical Stimulation Mapping (ESM) remains the clinical gold standard, it is invasive and time-consuming. To address this, we analyzed intracranial electrocorticographic (ECoG) data from 16 participants performing speech tasks and developed machine learning models to directly predict if the brain region underneath each ECoG electrode is critical. Ground truth labels indicating speech arrest were derived independently from Electrical Stimulation Mapping (ESM) and used to train classification models. Our framework integrates neural activity signals, anatomical region labels, and functional connectivity features to capture both local activity and network-level dynamics. We found that models combining region and connectivity features matched the performance of the full feature set, and outperformed models using either type alone. To classify each electrode, trial-level predictions were aggregated using an MLP applied to histogram-encoded scores. Our best-performing model, a trial-level RBF-kernel Support Vector Machine together with MLP-based aggregation, achieved strong accuracy on held-out participants (ROC-AUC: 0.87, PR-AUC: 0.57). These findings highlight the value of combining spatial and network information with non-linear modeling to improve functional mapping in presurgical evaluation.
\end{abstract}


\section{Introduction}
Mapping cortical regions essential for speech is critical to ensure the safety of brain surgeries involving language areas. This is particularly important in epilepsy surgery, where resection near language cortex may be required. Epilepsy affects over 65 million people worldwide ~\cite{thurman2017burden}, and many suffer from Drug-Resistant Epilepsy (DRE), where seizures persist despite treatment~\cite{kwan2010definition}. For these patients, surgical intervention may be the only option, but it carries risks to language and motor regions. Accurate pre-surgical mapping is therefore essential to avoid cognitive or functional impairments.

It is particularly challenging for localizing language function due to the complexity and individual variability. The clinical gold standard, Electrical Stimulation Mapping (ESM)~\cite{mandonnet2010direct}, identifies critical areas by directly stimulating the cortex during language tasks. However, ESM is invasive, time-consuming, and risky, potentially triggering seizures even in healthy tissue~\cite{borchers2012direct}.

Given the limitations of ESM, there is growing interest in less invasive alternatives for language mapping. Electrocorticography (ECoG) enables high-resolution recording of cortical activity via subdural electrodes during language tasks~\cite{miller2007real}. 

Many epilepsy patients already have ECoG grids implanted for monitoring and suppressing their seizures. Therefore, it is clinically feasible to ask patients performing a set of functionally relevant speech tasks similar to those performed during ESM before the surgery and  to use the recorded ECoG signals during these tasks to predict which electrodes correspond to speech-critical regions.

\section{Related Work}
While ECoG offers high spatial and temporal resolution, many conventional studies focus on signal amplitudes at individual electrodes, often overlooking broader spatial and network context~\cite{CRONE2001565}. This view can overlook the fact that language processing involves coordinated activities across distributed brain regions~\cite{fedorenko2014reworking, bassett2017network}.

More recently, Hsieh et al.~\cite{hsieh2024cortical} used functional connectivity and community structure to identify language-critical sites, focusing solely on network topology. 

To expand on this, our approach integrates connectivity with anatomical region information and neural activity features to provide a more comprehensive view of language-related brain function. We implement this in a machine learning framework that combines these complementary feature types to predict speech-critical cortical sites from ECoG data collected during a set of functionally relevant speech tasks.

\section{Data and Experimental Setup}

\subsection{Experimental Design}
This study involved 16 native English-speaking participants (9 female, 7 male, mean age = 29.48) with refractory epilepsy. Each participant was undergoing clinical monitoring at NYU Langone Hospital and had electrocorticography (ECoG) subdural electrode grids implanted as part of their standard care. The ECoG grid is  made up of standard 8 × 8 macro contacts spaced 10 mm apart (Ad-Tech Medical Instrument, Racine, WI). Electrode placement was based on clinical needs and covered key brain regions, including the superior temporal gyrus (STG), inferior frontal gyrus (IFG), and the pre- and postcentral gyri. 
The Institutional Review Board of NYU Grossman School of Medicine approved all experimental procedures. Written and oral consent was obtained from each participant following a detailed consultation with the clinical care team. The placement of ECoG electrodes was guided by clinical needs. Participants engaged in a set of functionally relevant tasks designed to elicit similar speech responses as Electrical Stimulation Mapping (ESM), but without applying stimulation. These tasks included:
\begin{itemize}
    \item \textbf{Auditory Word Repetition (AR)}: Repeating words heard audibly.
    \item \textbf{Auditory Word Naming (AN)}: Naming a word based on an auditory definition provided.
    \item \textbf{Sentence Completion (SC)}: Completing the final word of an auditory sentence.
    \item \textbf{Word Reading (WR)}: Reading aloud written words presented visually.
    \item \textbf{Picture Naming (PN)}: Naming a word based on a colored line drawing.
\end{itemize}

Each participant performed these tasks with a common set of 50 unique target words, which appeared once in the AN and SC tasks and twice in the remaining tasks, resulting in a total of 400 trials per participant ~\cite{shum2020neural}. 

\subsection{Data Collection and Preprocessing}

We analyzed ECoG recordings collected during functional speech tasks, separate from the stimulation mapping period. Electrodes with visible artifacts (e.g., line noise or poor contact) were excluded from all subsequent analyzes. Signals were re-referenced using the Common Average Reference (CAR) by subtracting the mean across all electrodes from each channel. Continuous data were segmented into 750 ms epochs time-locked to speech onset (250 ms before to 500 ms after articulation). Each trial was baseline-corrected and normalized using the mean activity during the 250 ms pre-stimulus period.

Our primary analysis focused on \textbf{High-Gamma Analytic Amplitude} (70–150 Hz), obtained by bandpass filtering the CAR-referenced signal and applying the Hilbert transform to extract the analytic envelope. For ablation studies, the raw CAR-referenced voltage was used directly, preserving low-frequency components and broader local field potentials. All preprocessing was performed using MATLAB and Python.

\subsection{Electrode Labeling Strategy During ESM}
Ground truth labels were derived from Electrocortical Stimulation Mapping (ESM) where clinicians identified electrodes as critical if stimulation reliably induced speech arrest, defined as an inability to speak without motor impairment (i.e., the patient can move but cannot speak)~\cite{kabakoff2024timing}. ESM stimulates electrode pairs; if speech arrest occurred, both electrodes in the pair were labeled as critical. Electrodes tested without arrest were labeled non-critical, and untested electrodes were excluded to ensure label reliability.

\section{Methods}

\subsection{Classification Framework}
\begin{figure*}[htbp] 
    \centering
    \includegraphics[width=0.6\textwidth]{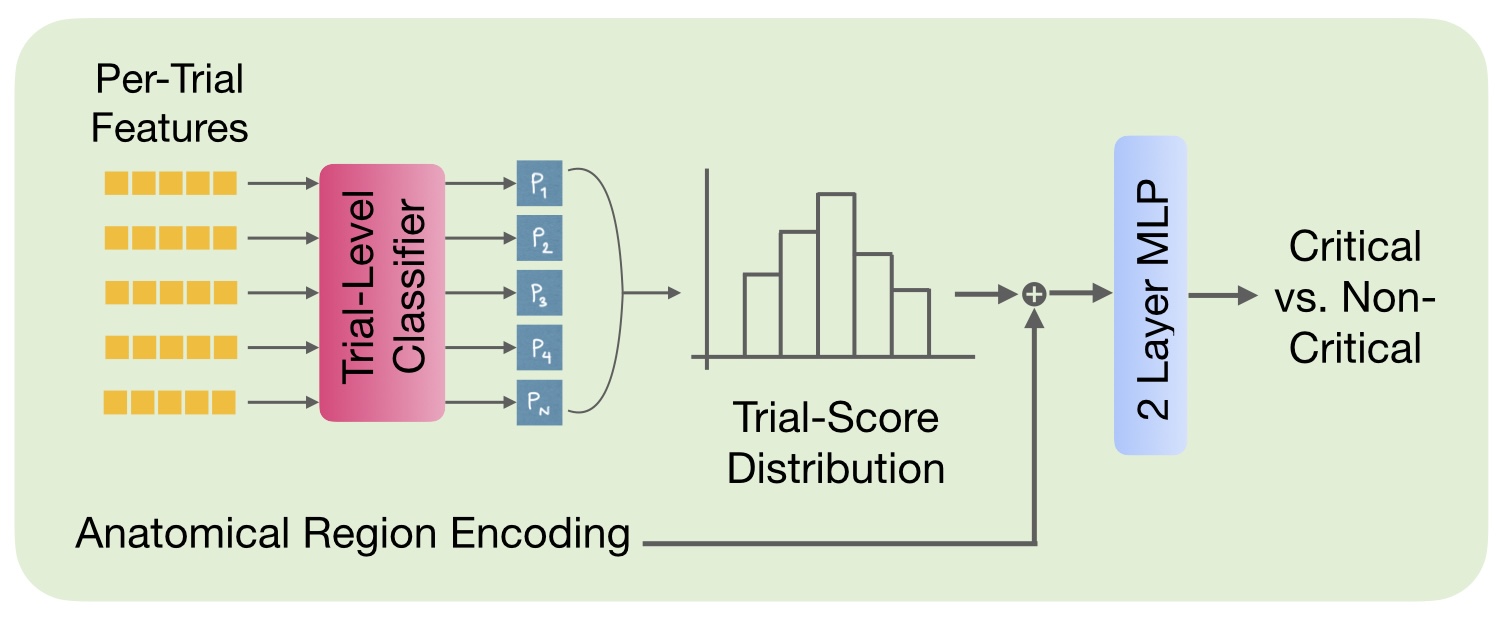}
    \caption{Two-stage classification pipeline with MLP-based aggregation of trial-level predictions.}
    \label{fig1}
\end{figure*}

The objective of the classification framework is to predict whether an electrode is \textit{critical} or \textit{non-critical} for speech production based on neural activity during speech tasks. Because we have neural data from multiple trials of each electrode, our classifier has two stages, as illustrated in Fig.~\ref{fig1}. The first stage applies a trial-level classifier, which takes the features extracted from the signal captured from each trial for an electrode and predicts the probability that the electrode is critical. The second stage takes the output probabilities from all trials and generates the final prediction for the electrode through another classifier.  Both classifiers are supervised. Next, we discuss the features and classifiers used in each stage.

\subsection{Feature Extraction for Trial-Level Classifier}

We use four different types of input features. These features were carefully selected to capture unique aspects of neural activation during task execution. These features are extracted from each test trial of the speech tasks  and concatenated to form the final input vector for the trial-level classifier.

\subsubsection{\textbf{NMF Feature Correlations}}
Instead of using the ECoG signal directly, which has a high dimension, we identify prototypical signal patterns (called prototype components) and represent each signal as a weighted average of these components. We identify the prototype components through  
Non-negative Matrix Factorization (NMF), which factorizes a non-negative data matrix \(X\) into two lower-dimensional non-negative matrices: a basis matrix \(T\) and a coefficient matrix \(H\), such that  $X \approx T H$

where:
\begin{itemize}
    \item \(X \in \mathbb{R}^{M \times N}\) is the original data matrix, where \(M\) is the number of samples in each trial, and \(N\) is the total number of trials across all electrodes and all subjects in the training data,
    \item \(T \in \mathbb{R}^{M \times K}\) represents the learned components (basis matrix), and $K$ is the number of prototype components,
    \item \(H \in \mathbb{R}^{K \times N}\) encodes the contribution of each component (coefficient matrix).
\end{itemize}

In our framework, ECoG signals from the full speech production window (-250 ms to +500 ms) for each trial were collected across all valid trials and electrodes and assembled into a matrix of shape \([M, N]\), where each column represents the signal from one electrode-trial pair. 

We applied NMF with $K=5$ components, yielding a set of prototype temporal patterns (the matrix T). Correlation scores between each trial and these five components were then computed to form the NMF-based features used for classification.

\subsubsection{\textbf{Mean Neural Activity}}
We compute the temporal mean of the ECoG signal  for each electrode-trial pair during the production window. This feature captures the mean neural activity throughout the pre-articulation and articulation time window, offering a representation of local neuronal engagement linked to speech processing.

\subsubsection{\textbf{Anatomical Region Encoding}}
Each electrode was assigned to one of 26 anatomical regions based on DKT Atlas~\cite{klein2012101}. These regions were encoded as a \textit{one-hot vector} of length 26. This encoding allows the classifier to be spatially aware of the electrode's anatomical context, which is crucial given that different brain regions have specialized functions.

\subsubsection{\textbf{Graph-Based Connectivity Features}}
To capture interactions between electrodes of the same subject, we computed functional connectivity using the \textit{Pearson correlation coefficient} for each pair of electrodes in the same test trial. We then took the absolute value of the correlation coefficients, resulting in a symmetric, undirected, non-negative connectivity matrix  $W=w_{i,j}$ for each trial, where $w_{ij} $ is the absolute Pearson correlation between nodes \(i\) and \(j\) for a single trial. Figure 2 illustrates an example of such a connectivity matrix.

\begin{figure}[htbp]
\centering
\includegraphics[width=0.25\textwidth]{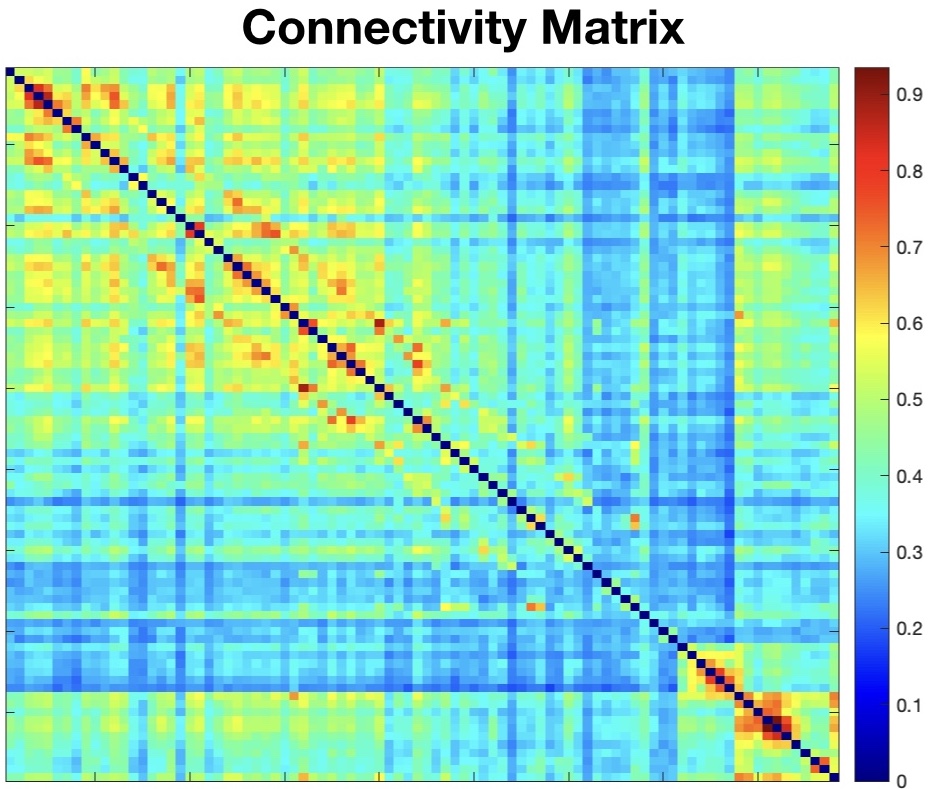}
\caption{Example connectivity matrix from one subject during a Word Reading trial. Rows and columns represent electrodes, and color intensity reflects functional connectivity strength, with warmer colors indicating stronger connections.}
\label{fig:connectivity_matrix}
\end{figure}

Next, these connectivity matrices were converted into weighted graphs, with electrodes as nodes and edge weights as pairwise correlations. This transformation allowed us to leverage graph-theoretical measures for a deeper analysis of network properties, following established approaches in network neuroscience~\cite{bullmore2009complex, rubinov2010complex, zuo2012network}. Specifically, we extracted the following features for each electrode:

\paragraph{\textbf{Strength}}  
The \textit{strength} of an electrode quantifies its overall connectivity by computing the mean of the absolute Pearson correlation coefficients with all other electrodes:
\begin{equation}
    S(i) = \frac{1}{L-1} \sum_{j\neq i} w_{ij}.
\end{equation}
where \(L\) is the total number of electrodes per subject.

This metric reflects the overall influence of an electrode in the network, capturing both its inflow and outflow interactions.

\paragraph{\textbf{Eigenvector Centrality (EC)}} 
\textit{Eigenvector centrality} measures the influence of an electrode based not only on its direct connections but also on the connectivity of its neighbors. Electrodes connected to other highly influential electrodes receive higher centrality scores. To compute the EC, we compute the eigenvector ${\bf e}$ corresponding to the largest eigenvalue  $\lambda$ of the adjacency matrix. The eigenvector centrality $E_i$ for an electrode \(i\) is simply the $i$-th component of this eigenvector, which satisfies:
\begin{equation}
    E_i = \frac{1}{\lambda} \sum_{j} w_{ij} E_j
\end{equation}

Electrodes with high eigenvector centrality are not only well-connected themselves, but are also connected to other highly central nodes, amplifying their network influence.

\paragraph{\textbf{Clustering Coefficient (CC)}}
The \textit{clustering coefficient} of an electrode measures the degree to which its neighboring electrodes form a tightly-knit cluster. It quantifies the local connectivity density around each node. The clustering coefficient for a node \(i\) is calculated as:
\begin{equation}
    C_i = \frac{\sum_{j,k} \left( w_{ij} w_{ik} w_{jk} \right)^{1/3}}{k_i (k_i - 1)}
\end{equation}
where:  
\begin{itemize}
    \item \( k_i \) is the the number of nodes connected to node $i$. Since we do not apply thresholding on the connectivity, \( k_i = L - 1 \), 
    \item The numerator sums over all possible pairs of neighbors \(j\) and \(k\) passing through $i$, accounting for their connection strengths.
\end{itemize}
A node with a high clustering coefficient indicates that its neighbors  are also well-connected with each other, suggesting strong local communication.

Finally, all extracted features  (NMF correlations, averaged neural activity, anatomical region encoding, and connectivity features) were concatenated into a single feature vector per electrode-trial. This representation captures both local activity and network-level context, and serves as input to the trial-level classification models for predicting criticality.

The process of feature extraction and aggregation is illustrated in Figure 3.

\begin{figure}[htbp] 
    \centering
    \includegraphics[width=0.48\textwidth]{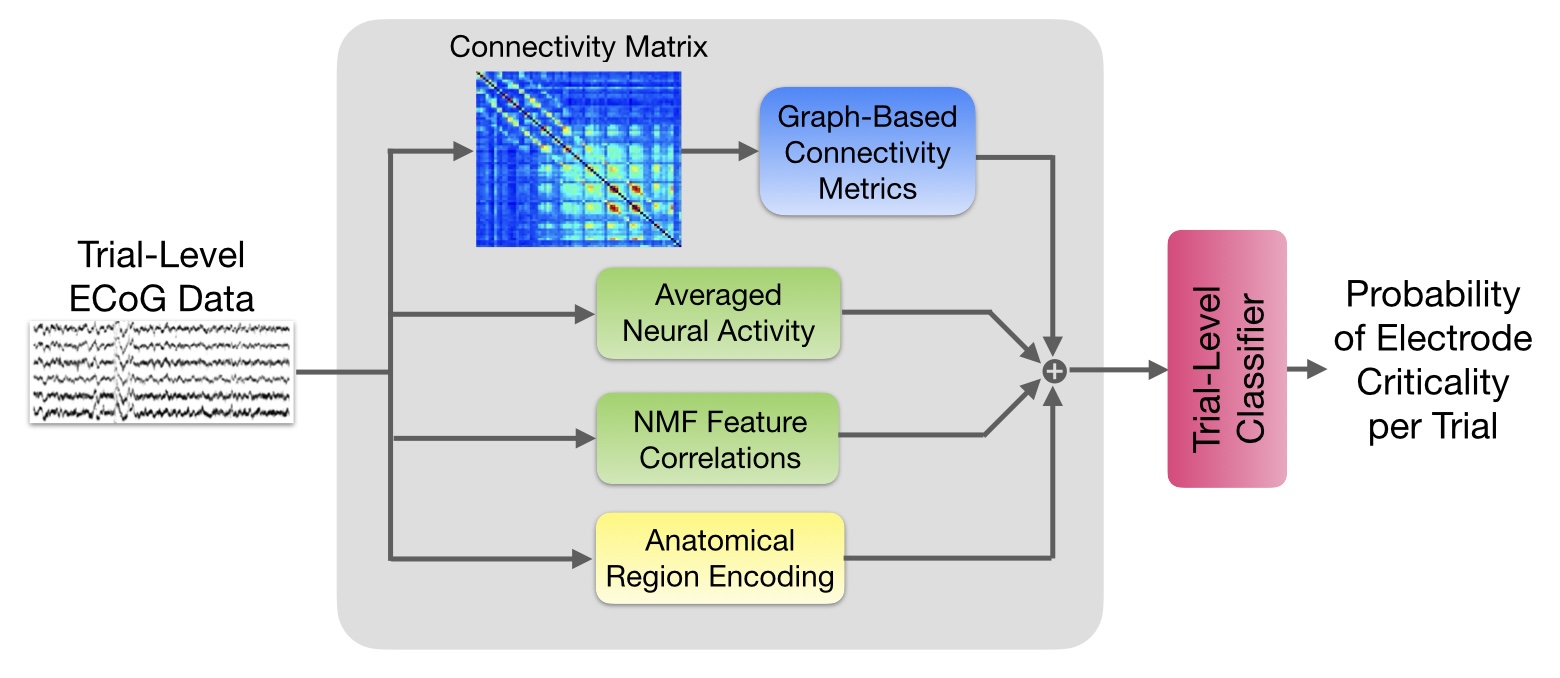}
    \caption{Feature aggregation pipeline combining neural, anatomical, and connectivity features for trial-level classification.}
    \label{fig3}
\end{figure}

\subsection{Trial-Level Classifier Models}
We compared  both linear and non-linear classifiers. We used a linear \textit{Support Vector Machine (SVM)} as a baseline for its strong generalization abilities and stability across diverse feature representations. Additionally, we conducted an ablation study with alternative classifiers, including \textit{Radial Basis Function (RBF) SVM, 2-layer Multi-Layer Perceptron (MLP), Logistic Regression, Random Forest,} and \textit{Decision Tree}, to evaluate their generalization and performance improvements.

\subsection{Aggregation of Trial Predictions:}
Since each electrode has multiple trial-level predictions, we use two different strategies to aggregate the prediction results from all trials for the same electrode:

\subsubsection{\textbf{Averaging-Based Aggregation}}
This approach simply uses the mean of the predicted probabilities across all trials as the predicted probability for each electrode. This simple yet effective method smooths out trial-level variability, capturing the overall prediction likelihood.  

\subsubsection{\textbf{MLP-Based Aggregation}}
While averaging offers a simple way to combine trial-level scores, it misses potentially informative patterns in their distribution. Specifically, there may be a nonlinear relationship between the shape of the score distribution across trials and the true likelihood that an electrode is critical. We use the histogram of trial-level scores to model their distribution and apply a 2-layer MLP classifier to generate final electrode-level predictions, though other classifiers could also be used within this framework.

\begin{itemize}
    \item \textbf{Trial-Level Score Histogram:} For each electrode, we group its trial-level prediction scores into a fixed-length histogram with 10 bins uniformly distributed between 0 and 1. The resulting histogram is a good approximation of the distribution of the trial-level prediction scores, allowing the MLP to learn adaptive decision rules based on subtle patterns that may be indicative of electrode criticality.

    \item \textbf{Region-Aware MLP Integration:} The histogram vector is then concatenated with the electrode's Region Information, a 26-dimensional one-hot encoding representing its anatomical location. This combined feature vector serves as the input to the MLP, which consists of an input layer accepting a 36-dimensional vector (10 histogram bins and 26 region encodings), followed by two hidden layers with 64 and 32 neurons activated by \texttt{ReLU}. The final output layer produces a single logit, passed through a \texttt{Sigmoid} activation to generate a probability score for classifying the electrode as \textit{critical} or \textit{non-critical}. Including region information allows the model to adapt its classification behavior based on both score distribution and anatomical priors (e.g., being more conservative or aggressive depending on the region’s typical involvement in language function). To address class imbalance between critical and non-critical electrodes (approximately 1:5 ratio), we trained the MLP using Focal Loss ~\cite{lin2017focal} with the Adam optimizer (learning rate = 0.001) for 20 epochs. 

\end{itemize}

\subsection{Evaluation}
\subsubsection{Validation Strategies}
We evaluated the model's performance using two complementary validation strategies:

\begin{itemize}
    \item \textbf{Electrode-Level Cross-Validation (CV):}  
    In this approach, electrodes are randomly partitioned into 8 folds regardless which subjects they belong to.   In each iteration, 7 folds are used for training and 1 for testing, cycling through all folds. This setup provides a robust estimate of model performance across diverse electrode samples, though it may include data from the same subject in both training and testing.

    \item \textbf{Leave-One-Subject-Out (LOO):}  
    To evaluate cross-subject generalization under more realistic clinical conditions, models are trained on all subjects except one, which is held out for testing. This procedure is repeated for every subject, ensuring that each evaluation is performed on completely unseen individuals.
\end{itemize}

\subsubsection{Metrics Reported}
To evaluate model performance, we report two key metrics:  

\begin{itemize}
    \item \textbf{AUC-ROC:}  
    Measures the model’s ability to distinguish critical from non-critical electrodes across thresholds. It calculates the area under the curve of Recall (the  True Positive Rate) vs. False Positive Rate, obtained by varying the threshold  on the predicted probability  for binarizing the prediction outcome.

    \item \textbf{AUC-PR:}  
    More informative under class imbalance, as it focuses on the model’s precision in identifying true positives.  It evaluates the area under the curve of the Precision vs. Recall. In our case, where critical electrodes account for approximately 17\% of the dataset, AUC-PR provides a more sensitive evaluation of model performance on the minority class.

    \item \textbf{Single-threshold metrics:} We also report Accuracy, Precision, Recall, F1-score, and Balanced Accuracy. For each fold, the decision threshold was chosen to maximize F1-score on the training data.
\end{itemize}

\section{Results}
We first established a baseline model using all four feature types (NMF feature correlations, averaged neural activity, anatomical regions, and connectivity metrics) extracted from High-Gamma signals. Trained with a Linear SVM and averaging-based aggregation, this model achieved a AUC-PR of 0.55 and 0.41 under cross-validation and leave-one-out validation, respectively, and AUC-ROC scores of 0.85 and 0.77 (Table~\ref{tab:core_ablation_svm}). The following sections evaluate the impact of different components of the overall classifier. 

\subsection{Feature Contribution Analysis}
\begin{figure}[htbp] 
    \centering
    \includegraphics[width=0.49\textwidth]{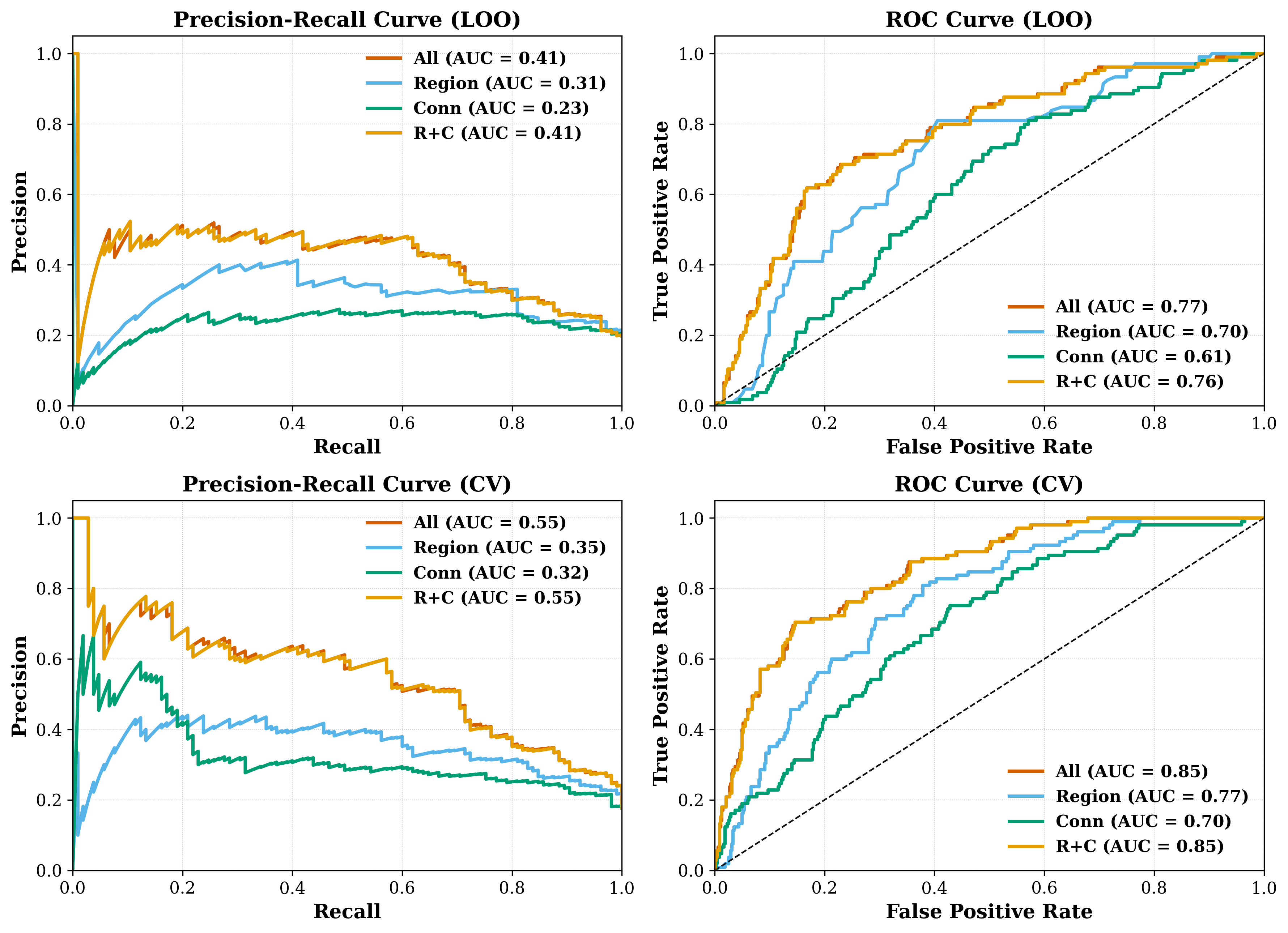}
    \caption{
    PR and ROC curves comparing feature set performance under Leave-One-Out (top row) and Cross-Validation (bottom row). Trial-Level Classifier: \textit{Linear SVM}; Signal: \textit{High-Gamma}; Aggregation: \textit{Averaging}.
    }
    \label{fig5}
\end{figure}

 We began by evaluating how different types of input features contribute to model performance. We compared the baseline model to variants using subsets of the input features, while keeping the signal type, classifier, and aggregation method fixed.  As shown in Table~\ref{tab:core_ablation_svm} and Figure~\ref{fig5}, combining region and connectivity features achieves performance nearly identical to using all features, indicating that these two sources of information carry the most predictive value.

\begin{table}[htbp]
    \centering
    \renewcommand{\arraystretch}{1.2}
    \setlength{\tabcolsep}{4pt}
    \caption{
    Results showing AUC scores (PR and ROC) under Cross-Validation and Leave-One-Out settings for different feature combinations. \textit{All Features} refers to the inclusion of all four feature types.
    }
    
    \begin{tabular}{lcccc}
        \hline
        \textbf{Configuration} & \textbf{PR-CV} & \textbf{PR-LOO} & \textbf{ROC-CV} & \textbf{ROC-LOO} \\ 
        \hline
        All Features & \textbf{0.55} & \textbf{0.41} & \textbf{0.85} & \textbf{0.77} \\ 
        Region + Connectivity & \textbf{0.55} & \textbf{0.41} & \textbf{0.85} & \textbf{0.76} \\ 
        Region Only & 0.35 & 0.31 & 0.77 & 0.70 \\ 
        Connectivity Only & 0.32 & 0.23 & 0.70 & 0.61 \\ 
        \hline
    \end{tabular}
    \label{tab:core_ablation_svm}
\end{table}

When used individually, region features outperform connectivity features, especially in LOO validation, but both are substantially less effective in isolation.

It is important to note that the modest AUC-PR scores are because of the class imbalance in our dataset, where only 17\% of electrodes are critical. In such a situation, it is very hard to achieve both high recall and high precision.

Additional single-threshold metrics (Table~\ref{tab:ablation_binary}) confirm this pattern: Region + Connectivity features nearly match the full-model performance, with F1 scores of 0.533 and 0.535, respectively. These results underscore the complementary nature of spatial and network-level information in identifying critical electrodes. The Wilcoxon signed-rank test also confirmed that using both connectivity and region features produces statistically significant improvements over using only connectivity or only region information. 

\begin{table}[htbp]
    \centering
    \caption{
    Single-threshold classification metrics across feature sets, evaluated using Leave-One-Out settings.\\
    Metrics: \textbf{Acc.} = Accuracy, \textbf{Prec.} = Precision, \textbf{Rec.} = Recall, \textbf{F1} = F1-score, \textbf{Bal. Acc.} = Balanced Accuracy. }

    \begin{tabular}{lccccc}
        \hline
        \textbf{Config.} & \textbf{Acc.} & \textbf{Prec.} & \textbf{Rec.} & \textbf{F1} & \textbf{Bal. Acc.} \\
        \hline
        All & \textbf{0.786} & \textbf{0.471} & \textbf{0.619} & \textbf{0.535} & \textbf{0.723} \\
        Region + Conn. & 0.784 & 0.468 & 0.619 & 0.533 & 0.722 \\
        Region Only & 0.665 & 0.314 & 0.581 & 0.408 & 0.634 \\
        Conn. Only & 0.597 & 0.267 & 0.590 & 0.368 & 0.595 \\
        \hline
    \end{tabular}
    \label{tab:ablation_binary}
\end{table}

\subsection{Aggregation Method Comparison}

We compared trial-level aggregation strategies using High-Gamma signals, Region and Connectivity features, and a Linear SVM. While simple averaging performs well, MLP-based aggregation methods yield consistent improvements, particularly in LOO validation. Incorporating histogram features improves performance modestly, but combining histogram and region information achieves the best overall results (see Table~\ref{tab:aggregation_methods}). These findings highlight the value of capturing distributional patterns and spatial context when aggregating trial-level predictions.

\begin{table}[htbp]
    \centering
    \renewcommand{\arraystretch}{1.2}
    \setlength{\tabcolsep}{4pt}
    \caption{
    Comparison of aggregation methods using High-Gamma signals, Region and Connectivity features, and a Linear SVM as a trial-level classifier. 
    }

    \begin{tabular}{lcccc}
        \hline
        \textbf{Aggregation Method} & \textbf{PR-CV} & \textbf{PR-LOO} & \textbf{ROC-CV} & \textbf{ROC-LOO} \\ 
        \hline
        Averaging & 0.55 & 0.41 & 0.85 & 0.76 \\ 
        MLP (Hist Only) & 0.55 & 0.49 & 0.85 & 0.78 \\ 
        MLP (Hist + Region) & \textbf{0.62} & \textbf{0.55} & \textbf{0.87} & \textbf{0.84} \\ 
        \hline
    \end{tabular}
    \label{tab:aggregation_methods}
\end{table}

\subsection{Effect of Signal Type: Raw vs. High-Gamma}

We compared classification performance using raw ECoG signals versus High-Gamma signals (70–150 Hz) as input. Both models used the same features, classifier, and aggregation method. As shown in Table~\ref{tab:signal_variation}, High-Gamma consistently outperformed raw signals across all evaluation metrics, suggesting it may provide a more informative representation of task-relevant neural activity.

\begin{table}[htbp]
    \centering
    \renewcommand{\arraystretch}{1.2}
    \setlength{\tabcolsep}{4pt}
    \caption{
    Comparison of classification performance using connectivity features derived from Raw vs. High-Gamma signals. All models use Region and Connectivity features, a Linear SVM, and averaging-based aggregation.
    }

    \begin{tabular}{lcccc}
        \hline
        \textbf{Configuration} & \textbf{PR-CV} & \textbf{PR-LOO} & \textbf{ROC-CV} & \textbf{ROC-LOO} \\ 
        \hline
        High-Gamma Signal & \textbf{0.55} & \textbf{0.41} & \textbf{0.85} & \textbf{0.76} \\ 
        Raw Signal & 0.51 & 0.40 & 0.83 & 0.74 \\ 
        \hline
    \end{tabular}
    \label{tab:signal_variation}
\end{table}

\subsection{Classifier Comparison}

Finally, we evaluated several commonly used classifiers for the trial-level classifier under our best-performing setting: High-Gamma signals, Region and Connectivity features, and MLP-based aggregation (Table~\ref{tab:classifier_comparison}). The RBF-kernel SVM achieved the strongest overall performance, with the highest AUC scores in both validation settings. The 2-layer MLP also performed well, achieving the highest PR-CV score (0.64) and competitive ROC values. Linear SVM and Logistic Regression produced results that suggest linear models remain effective in this context. In contrast, tree-based classifiers showed lower generalization, particularly in the LOO setting.

\begin{table}[htbp]
    \centering
    \renewcommand{\arraystretch}{1.2}
    \setlength{\tabcolsep}{6pt} 
    \caption{
    Performance of different trial-level classifiers using High-Gamma signals, Region and Connectivity features, and MLP-based histogram aggregation.
    }
    \begin{tabular}{lcccc}
        \hline
        \textbf{Classifier} & \textbf{PR-CV} & \textbf{PR-LOO} & \textbf{ROC-CV} & \textbf{ROC-LOO} \\ 
        \hline
        \multirow{1}{*}{Linear SVM} & 0.62 & 0.55 & 0.87 & 0.84 \\ 
        \multirow{1}{*}{SVM RBF} & 0.63 & \textbf{0.57} & \textbf{0.89} & \textbf{0.87} \\ 
        \multirow{1}{*}{Logistic Regression} & 0.62 & 0.53 & 0.86 & 0.83 \\ 
        \multirow{1}{*}{2-Layer MLP} & \textbf{0.64} & 0.54 & 0.86 & 0.83 \\        
        \multirow{1}{*}{Random Forest} & 0.61 & 0.48 & 0.81 & 0.80 \\ 
        \multirow{1}{*}{Decision Tree} & 0.60 & 0.50 & 0.82 & 0.80 \\ 
        \hline
    \end{tabular}
    \label{tab:classifier_comparison}
\end{table}

Table~\ref{tab:best_model_metrics} further shows the performance of the final model, an RBF-kernel SVM using Region and Connectivity features at the trial level, followed by MLP-based histogram aggregation, evaluated with single-threshold classification metrics.

\begin{table}[htbp]
    \small
    \centering
    \caption{Performance of the best model}
    \setlength{\tabcolsep}{4.3 pt} 
    \renewcommand{\arraystretch}{1} 
    \begin{tabular}{lccccccc}
    \hline
    \textbf{Method} & \textbf{Acc.} & \textbf{Prec.} & \textbf{Rec.} & \textbf{F1} & \textbf{Bal. Acc.} & \textbf{PR} & \textbf{ROC} \\
    \hline
    LOO & 0.796 & 0.490 & 0.733 & 0.588 & 0.772 & 0.57 & 0.87 \\
    CV  & 0.853 & 0.570 & 0.695 & 0.627 & 0.791 & 0.63 & 0.89 \\
    \hline
    \end{tabular}
    \label{tab:best_model_metrics}
\end{table}

\section{Discussion}

This study demonstrates that combining anatomical region labels with functional connectivity provides a powerful and interpretable foundation for identifying speech-critical electrodes from ECoG data. We found that this combination consistently achieved performance on par with models using the full feature set, suggesting that spatial and network-level features are the most informative dimensions for this task. High-Gamma activity further improved performance over raw signals, reinforcing its utility as a marker of task-relevant neural dynamics during speech production. Additionally, our MLP-based aggregation method, which models score distributions across multiple trials for each electrode, improved generalization across participants, especially under Leave-One-Out validation. This suggests that the distribution of trial responses, not just their average, offers valuable insight into electrode criticality.

Together, these findings support a data-driven approach to presurgical language mapping that moves beyond isolated signals toward richer representations of neural activity and network integration. Our framework offers a promising direction towards augmenting traditional ESM procedures, especially when stimulation is not feasible.

\section*{Acknowledgment}
This work was supported by the National Science Foundation under Grant No. IIS-2309057 (Y.W., A.F.), and National Institute of Health R01NS109367, R01NS115929.

\bibliographystyle{IEEEtran}
\bibliography{refs}

\begin{thebibliography}{10}
\providecommand{\url}[1]{#1}
\csname url@samestyle\endcsname
\providecommand{\newblock}{\relax}
\providecommand{\bibinfo}[2]{#2}
\providecommand{\BIBentrySTDinterwordspacing}{\spaceskip=0pt\relax}
\providecommand{\BIBentryALTinterwordstretchfactor}{4}
\providecommand{\BIBentryALTinterwordspacing}{\spaceskip=\fontdimen2\font plus
\BIBentryALTinterwordstretchfactor\fontdimen3\font minus \fontdimen4\font\relax}
\providecommand{\BIBforeignlanguage}[2]{{%
\expandafter\ifx\csname l@#1\endcsname\relax
\typeout{** WARNING: IEEEtran.bst: No hyphenation pattern has been}%
\typeout{** loaded for the language `#1'. Using the pattern for}%
\typeout{** the default language instead.}%
\else
\language=\csname l@#1\endcsname
\fi
#2}}
\providecommand{\BIBdecl}{\relax}
\BIBdecl

\bibitem{thurman2017burden}
D.~J. Thurman, G.~Logroscino, E.~Beghi, W.~A. Hauser, D.~C. Hesdorffer, C.~R. Newton, F.~A. Scorza, J.~W. Sander, T.~Tomson, and E.~C. of~the International League Against~Epilepsy, ``The burden of premature mortality of epilepsy in high-income countries: a systematic review from the mortality task force of the international league against epilepsy,'' \emph{Epilepsia}, vol.~58, no.~1, pp. 17--26, 2017.

\bibitem{kwan2010definition}
P.~Kwan and M.~J. Brodie, ``Definition of refractory epilepsy: defining the indefinable?'' \emph{The Lancet Neurology}, vol.~9, no.~1, pp. 27--29, 2010.

\bibitem{mandonnet2010direct}
E.~Mandonnet, P.~A. Winkler, and H.~Duffau, ``Direct electrical stimulation as an input gate into brain functional networks: principles, advantages and limitations,'' \emph{Acta neurochirurgica}, vol. 152, pp. 185--193, 2010.

\bibitem{borchers2012direct}
S.~Borchers, M.~Himmelbach, N.~Logothetis, and H.-O. Karnath, ``Direct electrical stimulation of human cortex—the gold standard for mapping brain functions?'' \emph{Nature Reviews Neuroscience}, vol.~13, no.~1, pp. 63--70, 2012.

\bibitem{miller2007real}
K.~J. Miller, P.~Shenoy, J.~W. Miller, R.~P. Rao, J.~G. Ojemann \emph{et~al.}, ``Real-time functional brain mapping using electrocorticography,'' \emph{Neuroimage}, vol.~37, no.~2, pp. 504--507, 2007.

\bibitem{CRONE2001565}
\BIBentryALTinterwordspacing
N.~E. Crone, D.~Boatman, B.~Gordon, and L.~Hao, ``Induced electrocorticographic gamma activity during auditory perception,'' \emph{Clinical Neurophysiology}, vol. 112, no.~4, pp. 565--582, 2001. [Online]. Available: \url{https://www.sciencedirect.com/science/article/pii/S1388245700005459}
\BIBentrySTDinterwordspacing

\bibitem{fedorenko2014reworking}
E.~Fedorenko and S.~L. Thompson-Schill, ``Reworking the language network,'' \emph{Trends in cognitive sciences}, vol.~18, no.~3, pp. 120--126, 2014.

\bibitem{bassett2017network}
D.~S. Bassett and O.~Sporns, ``Network neuroscience,'' \emph{Nature neuroscience}, vol.~20, no.~3, pp. 353--364, 2017.

\bibitem{hsieh2024cortical}
J.~K. Hsieh, P.~R. Prakash, R.~D. Flint, Z.~Fitzgerald, E.~Mugler, Y.~Wang, N.~E. Crone, J.~W. Templer, J.~M. Rosenow, M.~C. Tate \emph{et~al.}, ``Cortical sites critical to language function act as connectors between language subnetworks,'' \emph{Nature communications}, vol.~15, no.~1, p. 7897, 2024.

\bibitem{shum2020neural}
J.~Shum, L.~Fanda, P.~Dugan, W.~K. Doyle, O.~Devinsky, and A.~Flinker, ``Neural correlates of sign language production revealed by electrocorticography,'' \emph{Neurology}, vol.~95, no.~21, pp. e2880--e2889, 2020.

\bibitem{kabakoff2024timing}
H.~Kabakoff, L.~Yu, D.~Friedman, P.~Dugan, W.~K. Doyle, O.~Devinsky, and A.~Flinker, ``Timing and location of speech errors induced by direct cortical stimulation,'' \emph{Brain Communications}, vol.~6, no.~2, p. fcae053, 2024.

\bibitem{klein2012101}
A.~Klein and J.~Tourville, ``101 labeled brain images and a consistent human cortical labeling protocol,'' \emph{Frontiers in neuroscience}, vol.~6, p. 171, 2012.

\bibitem{bullmore2009complex}
E.~Bullmore and O.~Sporns, ``Complex brain networks: graph theoretical analysis of structural and functional systems,'' \emph{Nature reviews neuroscience}, vol.~10, no.~3, pp. 186--198, 2009.

\bibitem{rubinov2010complex}
M.~Rubinov and O.~Sporns, ``Complex network measures of brain connectivity: uses and interpretations,'' \emph{Neuroimage}, vol.~52, no.~3, pp. 1059--1069, 2010.

\bibitem{zuo2012network}
X.-N. Zuo, R.~Ehmke, M.~Mennes, D.~Imperati, F.~X. Castellanos, O.~Sporns, and M.~P. Milham, ``Network centrality in the human functional connectome,'' \emph{Cerebral cortex}, vol.~22, no.~8, pp. 1862--1875, 2012.

\bibitem{lin2017focal}
T.-Y. Lin, P.~Goyal, R.~Girshick, K.~He, and P.~Doll{\'a}r, ``Focal loss for dense object detection,'' in \emph{Proceedings of the IEEE international conference on computer vision}, 2017, pp. 2980--2988.

\end{thebibliography}

\end{document}